\documentclass[conference]{IEEEtran}
\IEEEoverridecommandlockouts
\usepackage[
  letterpaper,
  top=57pt, left=48pt, right=48pt, bottom=43pt
]{geometry}

\usepackage{afterpage}
\usepackage{amsmath,amsfonts}
\usepackage{algorithmic}
\usepackage{array}
\usepackage[justification=centering]{caption}
\usepackage[caption=false,font=normalsize,labelfont=sf,textfont=sf]{subfig}
\usepackage{textcomp}
\usepackage{stfloats}
\usepackage{cite}
\usepackage{tabularx}
\usepackage{url}
\usepackage{verbatim}
\usepackage{xcolor}
\usepackage{hyperref}
\usepackage{graphicx}
\def\BibTeX{{\rm B\kern-.05em{\sc i\kern-.025em b}\kern-.08em
    T\kern-.1667em\lower.7ex\hbox{E}\kern-.125emX}}
\usepackage{balance}
\begin{document}
\newgeometry{top=60pt, left=48pt, right=48pt, bottom=43pt}

\afterpage{\restoregeometry}

\title{PRISM: \underline{P}ointcloud \underline{R}eintegrated \underline{I}nference via \underline{S}egmentation and Cross-attention for \underline{M}anipulation}
\author{Daqi Huang\textsuperscript{1}\IEEEauthorrefmark{1}, 
Zhehao Cai\textsuperscript{1}\IEEEauthorrefmark{1}, 
Yuzhi Hao\textsuperscript{1}, 
Zechen Li\textsuperscript{1}, 
Chee-Meng Chew\textsuperscript{1}\\
\textsuperscript{1}National University of Singapore\\\IEEEauthorrefmark{1}Equal contribution}

\twocolumn[{%
\renewcommand\twocolumn[1][]{#1}%
\maketitle
\begin{center}
    \centering
    \captionsetup{type=figure}
     \includegraphics[width=1.0\textwidth]{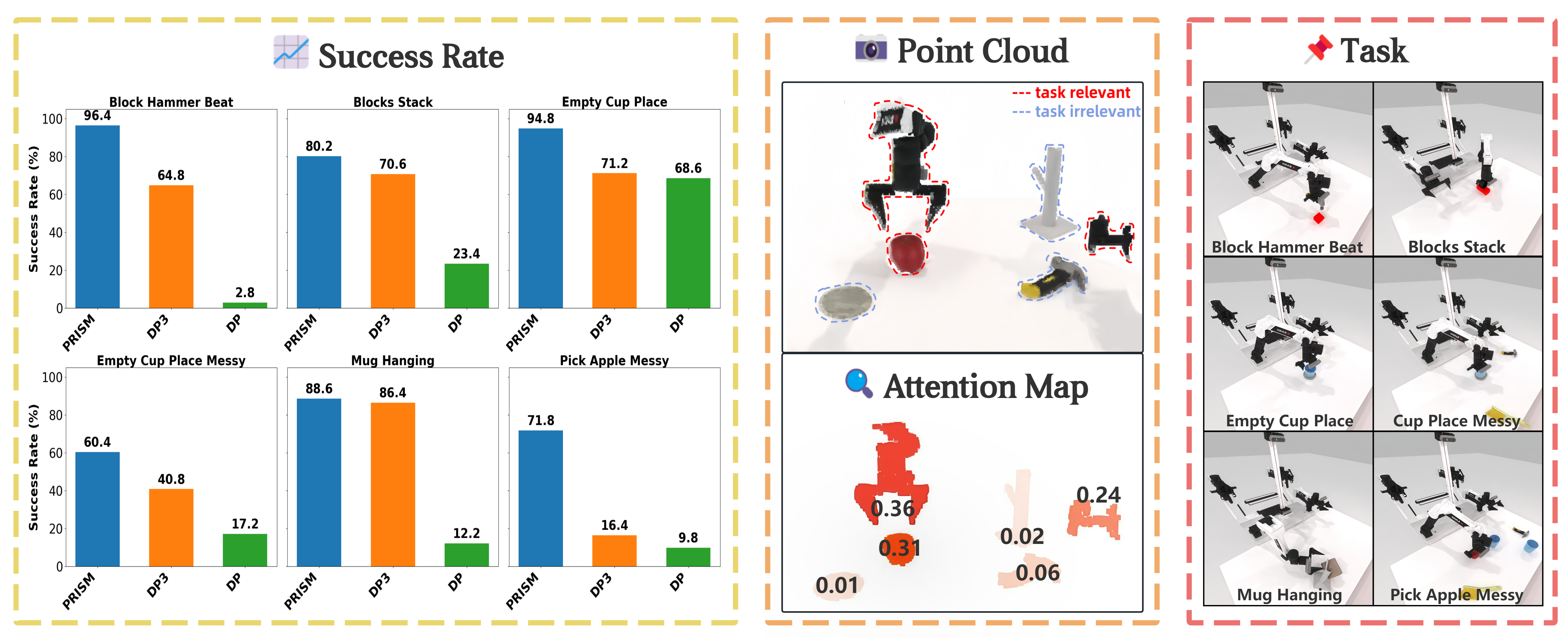}
    \caption{PRISM is a visual imitation learning algorithm that marries 3D visual representations with diffusion policies, achieving surprising effectiveness in diverse simulation and real-world tasks, with a practical inference speed.}
    \label{fig:teaser}
\end{center}

}]
\begin{abstract}
Robust imitation learning for robot manipulation requires comprehensive 3D perception, yet many existing methods struggle in cluttered environments. Fixed camera view approaches are vulnerable to perspective changes, and 3D point cloud techniques often limit themselves to keyframes predictions, reducing their efficacy in dynamic, contact-intensive tasks. To address these challenges, we propose PRISM, designed as an end-to-end framework that directly learns from raw point cloud observations and robot states, eliminating the need for pre-trained models or external datasets. PRISM comprises three main components: a segmentation embedding unit that partitions the raw point cloud into distinct object clusters and encodes local geometric details; a cross-attention component that merges these visual features with processed robot joint states to highlight relevant targets; and a diffusion module that translates the fused representation into smooth robot actions. With training on 100 demonstrations per task, PRISM surpasses both 2D and 3D baseline policies in accuracy and efficiency within our simulated environments, demonstrating strong robustness in complex, object-dense scenarios. Code and some demos are available on https://github.com/czknuaa/PRISM.
\end{abstract}

\IEEEpeerreviewmaketitle

\section{Introduction}
With advancements in robotics, the application scenarios for robotic arms are becoming increasely diverse . As robotic arms are required to interact with numerous objects in complex and dynamic environments, manipulation has emerged as one of the most cruicial aspects of the robotic systems\cite{zhang2025image,cai2024spatialbot,billard2019trends}. As robots are deployed in complex and variable settings such as vehicle assembly\cite{kang2020high} or healthcare support\cite{holland2021service}, advanced  manipulation skills are vital to achieving robust and optimal performance.

Learning-based methods have proven to be powerful tools for manipulation tasks due to their ability to handle task complexity These methods offer the  unique advantage of leveraging data directly to acquire target skills without the need for intricate manual tuning\cite{kroemer2021review,kleeberger2020survey}. Among the various approaches, imitation learning (IL) stands out as particularly effective because it circumvents the challenge of designing reward functions\cite{fang2019survey,zare2024survey}, a common hurdle in reinforcement learning(RL)\cite{liu2020reinforcement,hua2021learning,han2023survey}. Instead, IL learns directly from expert demonstrations, making it a practical solution for acquiring complex behaviors.

The workflow of an imitation learning (IL) model typically begins with collecting expert demonstrations, where an expert performs the desired task while their actions and corresponding states are recorded. This data is then preprocessed and transformed into a training dataset. Previous works have highlighted the effectiveness of IL methods, such as Behavior Cloning (BC)\cite{foster2024behavior,torabi2018behavioral,florence2022implicit}, Action Chunking with Transformers (ACT)\cite{zhao2023learning}, and Diffusion Policy (DP)\cite{chi2023diffusion}, across various manipulation tasks.
 However, these conventional approaches, which depend on processing holistic visual input, often struggle to deliver optimal policies in complex scene and cluttered environments.  This is because they fail to adequately focus on task-relevant features. 
 
 For instance, in a cluttered environment where a robot needs to pick a specific apple from a chaotic mix of fruits, utensils, and other objects, the complete scene input can obscure the target amidst irrelevant details, leading to confusion in policy generation\cite{mohammed2022review}. This challenge is even more pronounced in multi-stage tasks, where the robot must not only identify and grasp the target amidst the clutter but also execute subsequent actions, such as repositioning or stacking. Each phase demands focused attention on different aspects of the scene, a requirement that traditional holistic methods, which process the entire scene as a singular input, often find difficult to fulfill. For example, in a task where the robot must first pick up a cup and then place it on a mat, the model must shift focus between distinct elements of the task—an adaptability that conventional holistic approaches struggle to provide.

To address the challenges presented by both cluttered environments and multi-stage tasks, our approach, PRISM, utilizes point cloud segmentation as a fundamental first step. Instead of processing the entire scene as a single undifferentiated input—an approach that can obscure critical details—PRISM begins by decomposing the raw point cloud into distinct, object-centric clusters. This segmentation allows for the extraction of precise local features, even amidst distracting background elements.

Following this, a cross-attention mechanism merges these localized features with the robot’s joint state information.  This integration enables the system to dynamically focus on relevant targets for each phase of the task, whether it’s picking up an object from a complex scene or accurately placing it on a mat. By concentrating on pertinent details and minimizing interference from extraneous information, PRISM significantly enhances the model’s ability to adapt to the specific demands of both cluttered and multi-stage tasks.

In summary, our contributions are three-fold:
\begin{enumerate}
    \item We introduce a segmentation embedding module and a cross-attention module that allow  the model to autonomously focus on task-relevant objects, eliminating the need for  manual annotation.
\item We establish a cross-attention-based fusion mechanism that integrates segmented visual features with the robot’s joint state, achieving effective multi-modal fusion.
\item Our experimental results across several tasks show that PRISM consistently achieves high success rates, particularly in cluttered and multi-stage scenarios, underscoring its potential for effective and robust robotic manipulation.
\end{enumerate}


\section{Related Work}
\subsection{Vision-based Imitation Learning in Robotics}
Imitation learning aims to enable robots to replicate expert actions with high fidelity.  Traditional control methods often struggle with inaccuracies in object localization and the relative positioning of objects and robot arms\cite{ajwad2015systematic}. However, advancements in computer vison and depth camera have significantly mitigated these issues through enhanced visual observation capabilities. 

Imitation learning policies can be categorized based on the dimensionality of the input data: 2D image-based and 3D image-based approaches. Early research primarily focused on 2D policies, achieving notable success across various tasks. Algorithms such as Behavior Cloning (BC)\cite{foster2024behavior,torabi2018behavioral,florence2022implicit} and Generative Adversarial Imitation Learning (GAIL)\cite{ho2016generative} are commonly used to derive policies from demonstration data. BC leverages 2D data like object positions and hand trajectories to mimic expert behavior through supervised learning. While efficient, BC model often accumulate errors in scenarios lacking expert demonstration. GAIL addresses this limitation by incorporating adversarial training, reducing compounding errors and enabling robots to generalize across diverse visual scenarios. Similarly,  the Action Chunking with Transformer (ACT)\cite{zhao2023learning} framework effectively reduces error accumulation.

Compared to 2D-based methods, 3D imitation learning (3D-IL) approaches, which utilize point clouds, provide richer spatial information and depth perception. Techniques like ACT3D\cite{gervet2023act3d} and DP3\cite{ze20243d} have significantly outperformed the 2D model in multiple-task recoveries. For example, the DP3 method demonstrates a 24.2\% improvement in success rate over the Diffusion Policy (DP) method\cite{chi2023diffusion} across 72 tasks. However, models based on 3D input data encounter two primary challenges:

\textbf{(1) Indistinct input information.} The 3D models rely on point cloud data as input. The holistic nature of point cloud images can hinder the model’s ability to distinguish between different objects, particularly in complex scenes with multiple items.

\textbf{(2) Lack of multi-modal fusion.} In addition to vision point cloud
data, the robot’s joint state is crucial for the model. Current approaches often directly concatenate these features, which can lead to ineffective fusion due to inherent differences in representation and scale.

Recognizing these challenges, this paper introduces a novel architecture designed to segment input data and effectively fuse joint state information with vision data, thereby enhancing model performance.
\subsection{Point Cloud Processing}
Point cloud-based methods have gained significant atten- tion in the field of robotic manipulation due to their ability to capture rich spatial and geometric information\cite{guo2020deep,zhang2019review}. Unlike traditional 2D image-based approaches, point clouds provide accurate depth information and detailed 3D struc- tures, effectively addressing the limitations of 2D data in representing depth and spatial relationships in real -world environments. This enables robots to achieve a more comprehensive understanding of complex environments.

To process the unordered point cloud data, researchers have proposed end-to-end deep learning architecture such as PointNet\cite{qi2017pointnet}, which embeds each point into a high-dimensional feature space using shared multilayer perceptrons and employs max pooling to extract global features, thereby addressing the permutation invariance problem inherent to point clouds. Building on this framework, PointNet++\cite{qi2017pointnet++} introduced multi-scale grouping and dynamic sampling to enhance local feature extraction. Furthermore, DGCNN\cite{wang2019dynamic} utilized dynamic graph construction and edge convolution to iteratively update adjacency relationships and capture local features, thereby improving the topological representation and feature learning capabilities of point clouds. These methods represent common techniques for point cloud feature extraction.

In this paper, we adopt an approach that first segments the point cloud through clustering and then processes the segmented point clouds individually using a PointNet-derived architecture with LayerNorm optimization. This effectively addresses the limitations of PointNet in capturing local features and enhances the model’s ability to better understand the environment.

\section{Method}
\noindent Our objective is to learn an end-to-end policy $\pi_\theta$ that maps raw multi-modal sequences of historical observations \( \mathcal{O}^t \) =  \( \{{O}^{t-m}, {O}^{t-m+1}, \dots, {O}^t\} \)  to a sequence of predicted actions  \( \mathcal{A}^t \)=\( \{{A}^{t+1}, {A}^{t+2}, \dots, {A}^{t+n}\} \) through imitation learning from human demonstration. Each observation \( {O}^t = \{{P}^t, {S}^t\} \) consists of the 3D point cloud \( {P}^t \subset \mathbb{R}^{N \times 3} \) and the robot state \( {S}^t \in \mathbb{R}^{2 \times 7} \). Unlike previous methods that rely on pre-trained 3D representations or auxiliary tasks, PRISM is trained from scratch in an end-to-end fashion, directly mapping raw multi-modal sequences (point clouds + robot states) to actions using only the provided task demonstrations. Unlike conventional global point cloud feature extraction methods (\({F}^t_{\text{global}} = \Phi({P}^t) \)), PRISM enables the robot to attend to task-relevant objects autonomously through unsupervised scene decomposition and cross-modal attention. Specifically, raw point clouds \( {P}^t \) are segmented into \( K \) clusters \( \{{C}_k^t\} _{k=1}^K\) based on density without manual object labeling.Then, based on a cross-attention mechanism, the robot state features \({F}_s^t \) are used to query the point cloud features \( {F}_p^t \) to compute the attention scores \( \{\alpha_k\} _{k=1}^K\) of different objects and output the final conditioned features \( {F}_c^t \) . These features are subsequently fed into the Diffusion module to generate predicted actions.This approach suppresses irrelevant regions while adaptively allocating attention weights to critical objects based on their task relevance, effectively solving the attention dilution problem. 
\begin{figure*}[htbp]
    \centering
    \includegraphics[width=1\textwidth]{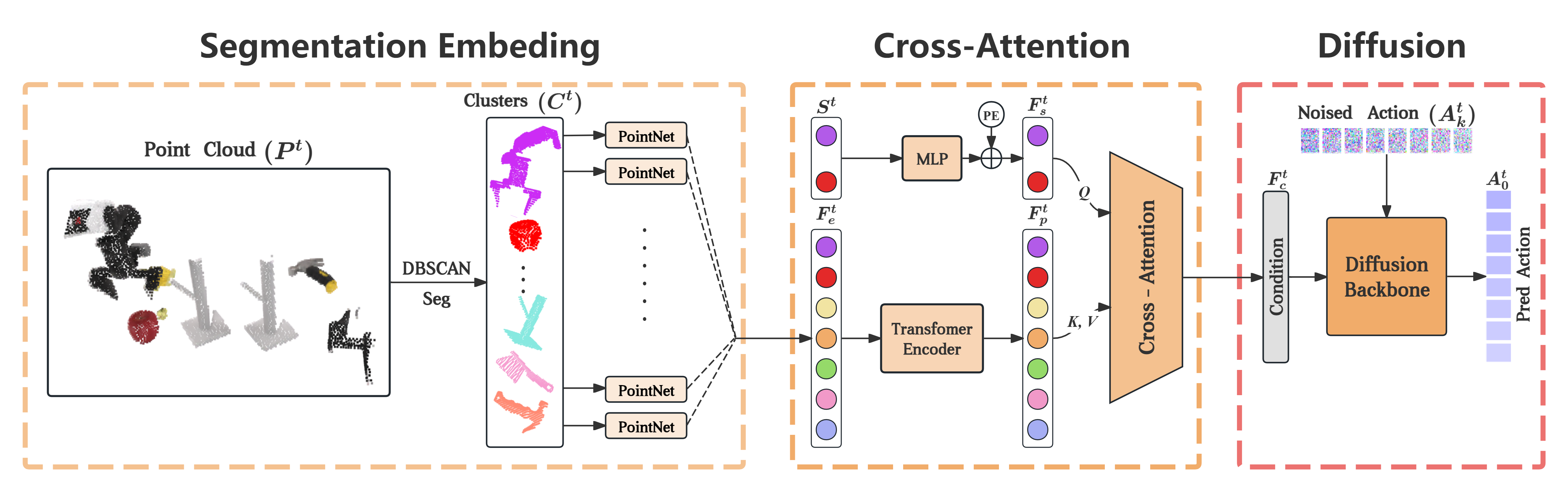}
    \caption{Over view of PRISM architecture. PRISM takes a single-view 3D point cloud as input. First, it segments the point cloud into clusters using DBSCAN, and each cluster is processed by a shared-parameter PointNet (batchnorm layer changed to layernorm layer) to extract local features. These features are fed into a Transformer encoder to capture inter-object relationships via self-attention. Next, the features are combined with the robot's proprioceptive state through cross-attention, focusing on task-relevant objects and generating conditional inputs for the diffusion module. Finally, the diffusion model denoises Gaussian noise step-by-step, guided by these conditions, to generate smooth and continuous action sequences.}
   
    \label{arc}
\end{figure*}

\subsection{Segmentation Embedding for Point Clouds}

\noindent The point cloud segmentation and feature extraction pipeline in PRISM functions similarly to the embedding layer in Transformer architectures. Unlike global max pooling or mean pooling operations that can dilute local details of objects , PRISM retains local features within clusters so that distinct object geometries and pose information are mapped to separate embedding subspaces. 

\textbf{Segmentation:} The raw 3D point cloud \( P^t = \{p_i^t\}_{i=1}^N \subset \mathbb{R}^{N\times3 }\) is divided into \( K \) object-centric clusters \(C^t = \{C_i^t \in \mathbb{R}^{N_k \times 3} \}_{k=1}^K\) using Density-Based Spatial Clustering of Applications with Noise(DBSCAN)\cite{ester1996density,schubert2017dbscan,mcinnes2017hdbscan}.
DBSCAN is a density-based clustering algorithm that identifies clustering structures through density connectivity. It categorizes points as: core points, border points, and noise points. A core point is a point whose \(\varepsilon\)-neighborhood contains at least \(minPts\) points(minPts is a hyperparameter); a border point is a point whose \(\varepsilon\)-neighborhood contains fewer than \(minPts\) points, but can be density-reachable from a core point; a noise point is a point that is neither a core point nor a border point.The clustering results of DBSCAN are optimized through the following stability criterion:
\begin{equation}
S(C^t_k) = \sum_{p_i^t \in C^t_k} \left( \lambda_{\text{core}}(p^t_i) - \lambda_{\text{noise}} \right)
\end{equation}
where \(\lambda_{\text{core}}(p_i)\) indicates whether point \(p_i\) is a core point, and  \(\lambda_{\text{noise}}\) is the density threshold for noise points. By maximizing this stability criterion, DBSCAN can generate high-quality clustering results.

The advantage of DBSCAN is that it does not require the number of clusters to be specified in advance and is an unsupervised method, eliminating the need for additional labeled data. In addition, a unified noise threshold \(\lambda_{\text{noise}}\) and minPts are used in all experiments, without the need to adjust parameters for different tasks.

While DBSCAN does not use semantic labels, its density-driven clustering eliminates the need for manual annotation, extensive pre-training, or large datasets—issues common to deep learning methods (e.g., 3D U-Net\cite{zhang20223d}, PointNet++\cite{qi2017pointnet++}) that rely on costly labeled data and slow training. Unlike these approaches, DBSCAN clusters objects *on-the-fly* using only geometric proximity, making it lightweight, computationally efficient, and ideal for robots. It naturally groups tightly positioned items (e.g., adjacent tools or overlapping parts) into cohesive units, critical for simultaneous multi-object interaction in cluttered environments.

Thereby significantly enhancing the universality and robustness of the algorithm.

\textbf{Embedding:} 
For each point cloud cluster \( C_k^t \), we extract its local features \( F_{ek}^t \in \mathbb{R}^{d_e} \) using a shared-parameter PointNet (batchnorm layer replaced by layernorm layer) \( \Phi_{\text{local}} \)\cite{qi2017pointnet}. PointNet operates directly on unordered point sets by using symmetric functions (max pooling) to achieve permutation invariance. Formally, given a cluster \( C_k^t = \{p_j^t\}_{i=1}^{N_k}\), the feature extraction process is:
\begin{equation}
F_{ek}^t = \Phi_{\text{local}}(C_k^t) = \text{MaxPool}\left\{ \text{MLP}(p_j^t) \mid p_j^t \in C_k^t \right\}
\end{equation}

where \(\text{MLP}\) denotes a multi-layer perceptron that maps each point \( p^t_j \) to a high-dimensional feature space, and MaxPool aggregates point-wise features into a unified descriptor for the object. The shared-parameter design ensures consistent feature encoding across all clusters while reducing model complexity.Compared to directly performing cross-attention between all point clouds and joint angles, this approach reduces the number of tokens, thereby decreasing computational load and improving data sample utilization efficiency.

\subsection{Transfomer Encoder for Relational Features Reshaping}
\noindent In robotic manipulation tasks, objects often serve distinct semantic roles across different phases.Therefore, the robot must dynamically shift its attention on objects according to these phase-specific roles. While modified PointNet extracts isolated object-level features \( F_e^t = \{F_{ek}^t\}_{k=1}^K \), these features lack contextual relationships between objects, which are crucial for understanding task stages in manipulation scenarios. To address this, we employ a Transformer encoder \( T \) to model inter-object dependencies and reshape the feature space\cite{vaswani2017attention}.The Transformer operates on the set of object features \( F_e^t  \) through self-attention mechanisms\cite{bahdanau2014neural}:
\begin{equation}
F_{p}^t = \text{Softmax}\left( \frac{(F_e^t W_Q)(F_e^t W_K)^\top}{\sqrt{d}} \right) F_e^t W_V
\end{equation}
where \( W_Q, W_K, W_V \) are linear projections of the input features \( F_e^t \). This allows the model to dynamically compute attention weights between objects, capturing their spatial and semantic relationships. 
By incorporating relational context, the Transformer enables the robot to infer task phases and adapt its behavior accordingly. This capability is particularly valuable in multi-object environments, where the meaning of an object depends on the current task context. The output features \( Fp^t \in \mathbb{R}^{K \times d_p} \) encode both object-level geometry and inter-object relationships, providing a comprehensive representation for downstream action generation.Section IV will demonstrate its effectiveness through ablation studies.

\subsection{Cross Attention}
\noindent In PRISM, a cross-attention mechanism is employed to effectively integrate the proprioceptive states \( F_s^t  \) with the visual features \( F_p^t  \) from the Transformer encoder. During manipulation, each hand performs distinct tasks, necessitating focus on different objects. To enable selective attention, the arms are decoupled, allowing each to independently query point cloud features and compute attention weights, thus focusing on their relevant features:

\begin{equation}
F_{c}^t = \text{Softmax}\left( \frac{(F_s^t W_Q)(F_p^t W_K)^\top}{\sqrt{d}} \right) F_p^t W_V
\end{equation}
where \( W_Q, W_K, W_V \) are learnable projection matrices.To eliminate the potential ambiguity between the joint states of the left and right arms, we embed a set of specific positional embeddings \( E_i \) into the joint angle features of each arm\cite{ke2020rethinking}:

\begin{equation}
\tilde{F}_{si}^t = F_{si}^t + E_i, \quad i \in \{L, R\}
\end{equation}
These positional embeddings provide a unique identity for each arm, ensuring that the attention weights \( \alpha_k^i \) remain independent even when the joint configurations are perfectly symmetrical. This design effectively enhances the model's ability to distinguish between the features of the left and right arms, thereby providing more precise input information for subsequent motion planning and task execution.

\subsection{Diffusion for Action Generation}
\noindent The action generation is formulated as a conditional diffusion process, where the decoder \(D\) progressively denoises an initial Gaussian distribution \(\mathcal{N}(0, \sigma^2 I)\) into executable actions \(A_t\), guided by the fused visual-proprioceptive feature \(\mathcal{F}_c^t\)\cite{chi2023diffusion, ze20243d, ho2020denoising}. At denoising step \(k\), the denoise process is:

\begin{equation}
\mathcal{A}^{t}_{k-1} = \beta \left( \mathcal A_{k}^t - \delta \epsilon_{\theta}((\mathcal{F}_c^t, \mathcal A_{k}^t, k) \right) + \mathcal{N}(0, \sigma^2 I)
\end{equation}

where \(\epsilon_{\theta}\) denotes a UNet-based\cite{siddique2021u} noise predictor parameterized by \(\theta\), with \(\beta\), \(\delta\),  being step-dependent coefficients from the DDIM noise scheduler \cite{song2021denoising,perez2018film}. The training objective minimizes the MSE loss between predicted and ground-truth noise:
\begin{equation}
L_{\text{diff}} = \mathbb{MSE} ( \epsilon_{\theta}(\mathcal{F}_c^t, \mathcal A_{k}^t, k) , \epsilon_k )
\end{equation}
where\(\mathcal{A}_{k}^{t}\)is the noised version of demonstration action \(A_{0}^t\) at diffusion step \(k\).


\section{Experiments}
To comprehensively validate PRISM’s performance under practical conditions, we investigate the following crucial questions:

\noindent\textbf{How robust is PRISM in cluttered, object-dense scenes?}  
We evaluate PRISM’s ability to pick out and manipulate a target object amid heavy visual interference by measuring its success rates in scenarios with many distractors.

\noindent\textbf{How effectively can PRISM complete multi-stage manipulation?}  
We assess its capacity to shift attention dynamically and carry out each sequential sub-goal—grasping, transporting, placing—in a single end-to-end pipeline.

\noindent\textbf{What is the impact of the transformer encoder in PRISM?}  
Through ablation studies across multiple tasks, we compare performance with and without the self-attention module to quantify its benefits and identify any overhead in low-data situations.

\noindent\textbf{How data-efficient is PRISM?}  
By varying the number of demonstrations (e.g., 20, 50, 100, 200), we chart PRISM’s learning curve against baseline policies to determine how quickly it attains high success rates.

\subsection{Experiment Setup}
\textbf{Simulation benchmark.} In this paper, benchmark from RoboTwin\cite{mu2024robotwin} is used for training and testing. 6 tasks are chosen from the benchmark in the experiment. These tasks span various challenging scenarios, including cluttered environment tasks and multi-stage tasks. All tasks are built on the Sapien simulator. The description of the 6 tasks used for experiment is shown in TABLE \ref{task}.

\begin{figure}[htbp]
    \centering
    \includegraphics[width=0.5\textwidth]{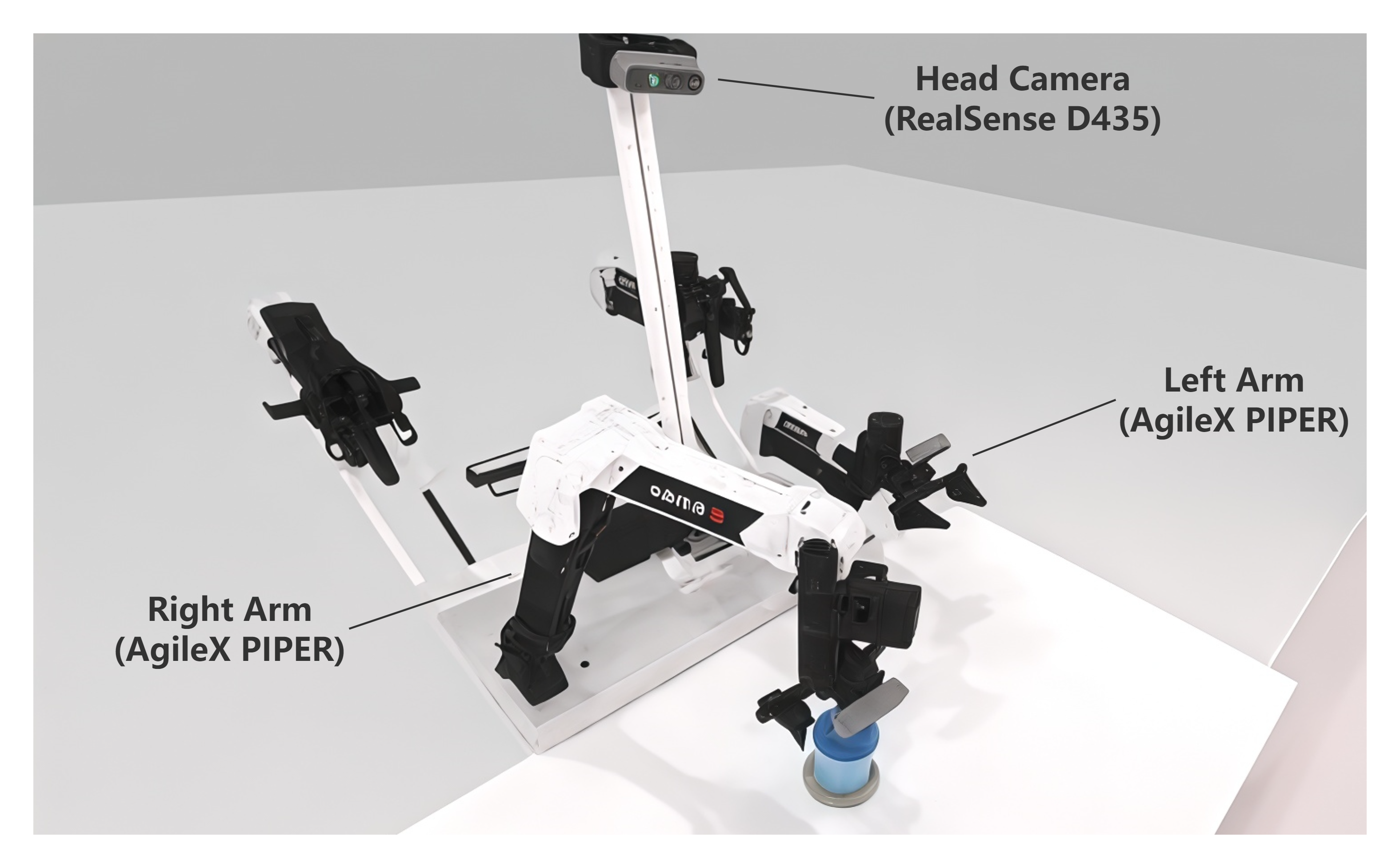}
    \caption{Cobot Magic platform in simulation environment\cite{mu2024robotwin}.}
    \label{cobot}
\end{figure}

\textbf{Environment setup.} The system integrates ManiSkill\cite{mu2021maniskill}—an open-source simulator offering GPU-accelerated data collection on the SAPIEN\cite{xiang2020sapien} framework—to simulate an open-source Cobot Magic platform with a Tracer chassis and four robotic arms, as depicted in Fig. \ref{cobot}. The rear two robotic arms of the platform are designed for teleoperation: no motion commands are input to these rear arms, nor do they output any information. Additionally, a single RealSense D435 camera is mounted at a high vantage point to capture a wide field of view, with data recorded at 15Hz. A single NVIDIA RTX Laptop 4060 GPU is used for data collection, training, and evaluation.

\textbf{Baselines.} Diffusion Policy \cite{chi2023diffusion} is a generative visuomotor framework that learns a distribution over actions; its 2D variant uses images or videos but lacks full 3D context. The 3D version \cite{ze20243d} encodes point clouds into compact features, improving performance when detailed spatial understanding is needed.

As PRISM is also diffusion-based method, we therefore employ 2D Diffusion Policy(DP)(resnet18+unet) and 3D Diffusion Policy(DP3)(xyz) as baselines in our subsequent experiments.

\textbf{Evaluation metrics.} Five different seed experiments (e.g., seeds 0, 1, 2, 3, and 4) are performed for each task, all tasks underwent 100 evaluation rollouts per policy configuration. During training, the model is saved at intervals of 200 epochs, generating 20 checkpoints per seed. For every seed, the checkpoint yielding the highest success rate is identified, and finally, the mean and standard deviation of these five peak success rates is computed and reported.
\begin{table*}[htbp]
    \centering
    \small 
    \caption{Task Descriptions}
    \label{task}
    \begin{tabularx}{\textwidth}{|l|X|}
    \hline
    \textbf{Task} & \textbf{Description} \\
    \hline
    \textit{Block Hammer Beat} (BHB) & A hammer and a block rest on the table’s center. If the block lies closer to the left manipulator, that side picks up the hammer and taps the block. Otherwise, the right manipulator carries out the same move. \\
    \hline
    \textit{Blocks Stack} (BS) & A rectangular cuboid and a cube are placed randomly on the table. The robotic arm first positions the rectangular piece in the designated target area, then stacks the cube on top, ensuring the proper order. \\
    \hline
    \textit{Empty Cup Place} (ECP) & A cup and its corresponding mat appear at arbitrary positions. The robot subsequently repositions the cup onto its assigned mat. \\
    \hline
    \textit{Empty Cup Place Messy} (ECPM) & A cup, its mat, and additional assorted objects are distributed unpredictably. The robot is tasked with repositioning the cup onto its designated mat. \\
    \hline
    \textit{Mug Hanging} (MH) & A mug is initially located on the table’s left side, while a rack is randomly placed on the right. One manipulator moves the mug toward the center, and the other arm then hangs it by its handle onto the rack. \\
    \hline
    \textit{Pick Apple Messy} (PAM) & Apples and various items are scattered across the workspace. The robot selects one apple from the clutter and lifts it. \\
    \hline
    \multicolumn{2}{|l|}{\footnotesize $^{a}$Descriptions are based on the experimental task setup.} \\
    \hline
    \end{tabularx}
\end{table*}

\begin{figure*}[htbp]
    \centering
    \includegraphics[width=1\textwidth]{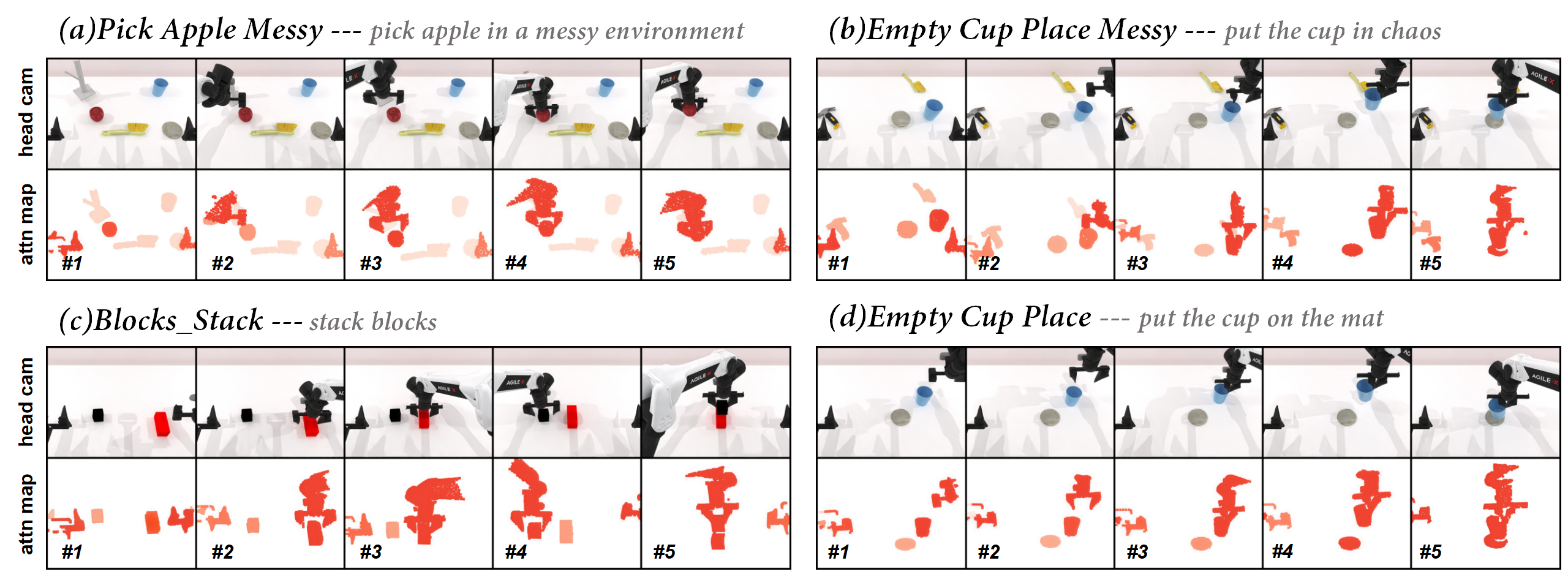}
    \caption{Attention heat map of Pick Apple Messy, Empty Cup Place Messy, Blocks Stack and Empty Cup Place tasks. Points are colored according to their attention weights using a gradient from the \texttt{matplotlib.cm} library. Darker shades indicate higher attention, while lighter hues reflect lower influence on the overall decision.}
    \label{atten}
\end{figure*}
\subsection{Cluttered Environment Tasks}
In this experiment, the cluttered environment tasks comprise two scenarios: Empty Cup Place Messy and Pick Apple Messy. These tasks are significant because they simulate real-world settings where objects are arranged in a disorderly and unpredictable manner. 

The cross attention mechanism, which computes interactions between joint state data and visual features, is able to pick out the target object from among many distractions. This selective focusing is also visually evident in the attention heat map shown in Fig. \ref{atten}(a) and (b), consider the Pick Apple Messy and Empty Cup Place task: although the workspace is cluttered with numerous items, only the target apple or cup draws significantly higher attention. This selective focusing demonstrates that the mechanism effectively filters out irrelevant clutter while emphasizing the essential object, thereby enhancing the system’s capability to perform reliably in unpredictable, messy environments.

In this experiment, PRISM consistently outperforms baseline methods. As shown in TABLE~\ref{success}, PRISM achieves an overall average success rate of 66.1\%, compared to 28.6\% and 13.5\% for DP3 and DP, respectively. Unlike the baselines—which process the entire scene as a single input and are often distracted by extraneous details—PRISM first segments the raw point cloud into object-centric clusters. This segmentation, combined with cross-attention that fuses these features with the robot's joint states, enables the model to focus exclusively on task-relevant objects. Consequently, PRISM exhibits significantly higher accuracy and robustness in challenging, cluttered scenarios.

\subsection{Multi-stage Tasks}
In this experiment, the multi-stage tasks include Block Hammer Beat, Blocks Stack, Empty Cup Place and Mug Hanging. These scenarios require a focus on distinct objectives in different phases of the operation. As each goal arises, the robot must adapt its planning and actions, showcasing its ability to handle evolving priorities and maintain situational awareness throughout the task.

The cross-attention mechanism between the joint state and visual features dynamically adjusts its focus according to the demonstration phase. For instance, as illustrated in the attention heat map in the Fig. \ref{atten}(d), the attention shifts from the cup to the mat after the grasping phase, validating the dynamic focus capability of the model during sequential task stages. In the Empty Cup Place task, the mat remains overlooked until the cup is grasped(before \#4); only afterward does the system shift its attention to the mat as the target for placement(after \#4). Take Blocks Stack task as another example: as depicted in Fig. \ref{atten}(c), the attention changes from the rectangular cuboid to the cube(after \#3) when the rectangular cuboid has been placed properly. This transition illustrates the mechanism’s capacity to reassign focus in real time, effectively differentiating between sequential sub-goals. Such adaptive attention is essential for executing multi-stage operations, where the ability to prioritize changing objectives significantly enhances task performance.

In the multi-stage task experiment, PRISM demonstrates superior performance over baseline approaches. As shown in TABLE \ref{success}, PRISM achieves an overall average success rate of 90.0\%, while DP3 and DP reach only 73.3\% and 26.8\%, respectively. Baseline methods typically treat the scene as a single, undivided entity, which makes it difficult for them to separate and address the sequential sub-goals inherent in multi-stage tasks. In contrast, our approach first decomposes the point cloud into discrete, object-focused clusters and then applies a cross-attention mechanism to combine these clusters with the robot’s joint state data. This targeted strategy enables the system to adapt its focus as different sub-goals emerge throughout the operation, allowing for dynamic reallocation of attention. Consequently, PRISM effectively isolates critical information at each stage, overcoming the limitations of holistic input processing and yielding markedly improved performance.

\subsection{Ablation Experiment on Transformer Encoder}
To highlight the essential role of the transformer encoder in the PRISM framework, ablation experiments were conducted with 100 demonstrations on six tasks: Block Hammer Beat, Blocks Stack, Empty Cup Place, Empty Cup Place Messy, Mug Hanging, and Pick Apple Messy. Specifically, the model was trained with and without the transformer encoder, and the corresponding success rates for each configuration are presented in Table~\ref{ablation}. The overall average success rate increases from 75.5\% to 82.0\% when the transformer encoder is incorporated. These improvements demonstrate that the self-attention mechanism effectively re-weights the most pertinent features, enabling the robot to focus more effectively on the target. As a result, the transformer encoder is crucial for enabling PRISM to robustly perceive and manipulate objects under challenging conditions.

\begin{figure}[htbp]
    \centering
    \includegraphics[width=1\linewidth]{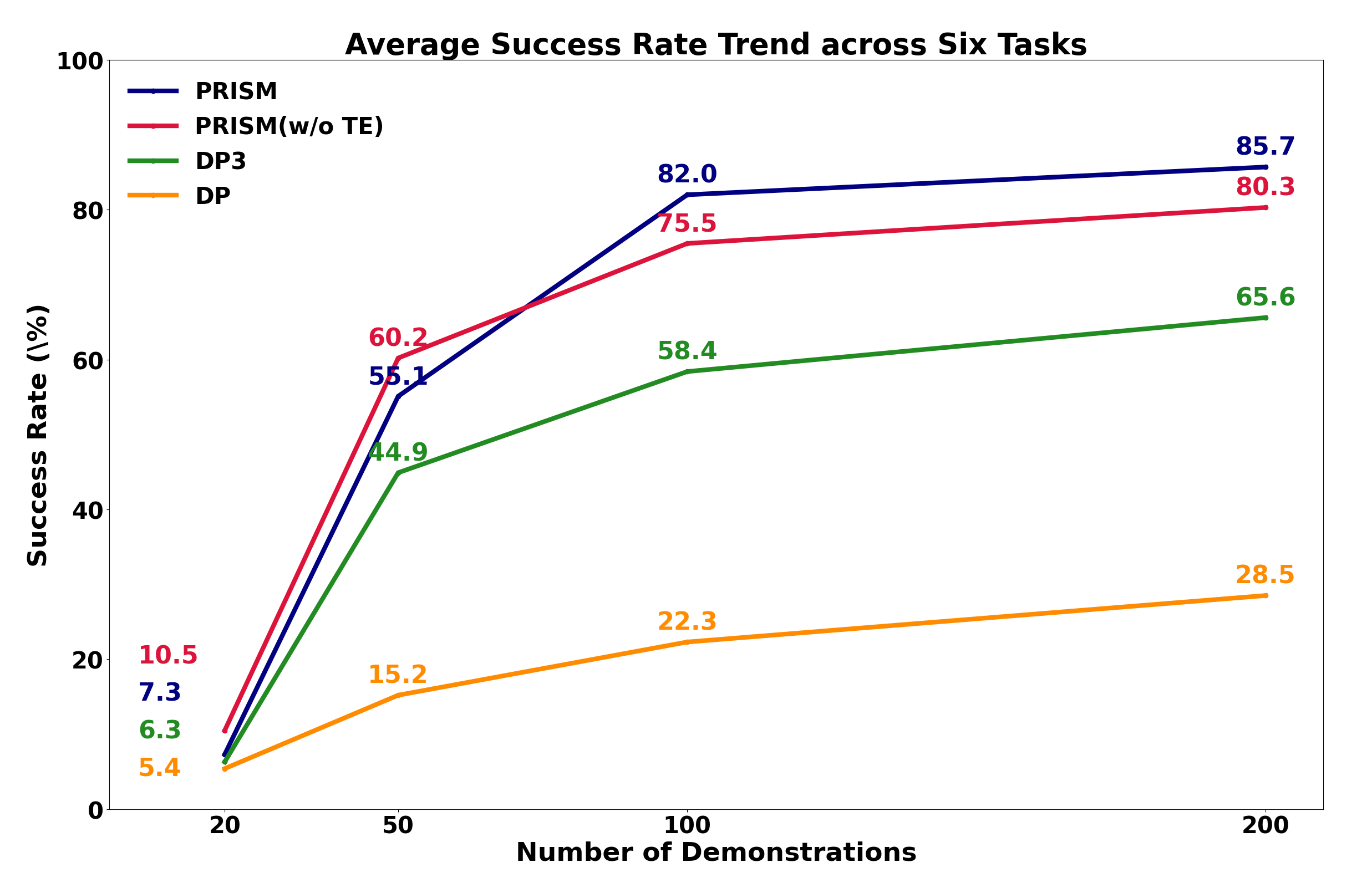}
    \caption{Success Rate Trend across 6 Tasks.}
    \label{zhe}
\end{figure}

However, further experiments, as depicted in Fig. \ref{zhe}, revealed some drawbacks associated with the encoder structure. While the self-attention mechanism adeptly highlights key features, it also introduces additional computational overhead. The large number of parameters can lead to overfitting when data is limited, suggesting that simpler architectures might be more effective in low-data settings. Therefore, although the encoder greatly improves performance, a lighter or more targeted attention mechanism could be more suitable for situations with limited data.

\begin{table}[htbp]
    \centering
    \caption{Ablation Experiment on Transformer Encoder}
    \label{ablation}
    \begin{tabular}{|l|c|c|}
    \hline
    \textbf{Task} & \textbf{PRISM (w/ TE)} & \textbf{PRISM (w/o TE)} \\
    \hline
    Block Hammer Beat    & 96.4 & 90.0 \\
    \hline
    Blocks Stack         & 80.2 & 75.2 \\
    \hline
    Empty Cup Place      & 94.8 & 87.8 \\
    \hline
    Empty Cup Place Messy& 60.4 & 55.4 \\
    \hline
    Mug Hanging          & 88.6 & 82.2 \\
    \hline
    Pick Apple Messy     & 71.8 & 62.6 \\
    \hline
    \textbf{Average}     & \textbf{82.0} & \textbf{75.5} \\
    \hline
    \end{tabular}
\end{table}

\begin{table}[htbp]
\centering
\caption{Success rates (mean ± std, \%) for models under 100 demos}
\label{success}
\begin{tabular}{|l|c|c|c|}
\hline
\multicolumn{1}{|c|}{\textbf{Task}} & \textbf{PRISM} & \textbf{DP3} & \textbf{DP} \\
\hline
\textbf{BHB} & 96.4 ± 1.52 & 64.8 ± 2.04 & 2.8 ± 0.84 \\
\hline
\textbf{BS} & 80.2 ± 0.84 & 70.6 ± 1.14 & 23.4 ± 1.67 \\
\hline
\textbf{ECP} & 94.8 ± 1.48 & 71.2 ± 0.84 & 68.6 ± 1.14 \\
\hline
\textbf{ECPM} & 60.4 ± 1.52 & 40.8 ± 0.84 & 17.2 ± 0.43 \\
\hline
\textbf{MH} & 88.6 ± 1.14 & 86.4 ± 0.89 & 12.2 ± 0.45 \\
\hline
\textbf{PAM} & 71.8 ± 1.48 & 16.4 ± 0.89 & 9.8 ± 0.84 \\
\hline
\textbf{Cluttered} & 66.1 ± 6.17 & 28.6 ± 12.89 & 13.5 ± 3.95 \\
\hline
\textbf{Multi-stage} & 90.0 ± 6.62 & 73.3 ± 8.30 & 26.8 ± 25.55 \\
\hline
\textbf{Overall} & 82.0 ± 13.11 & 58.4 ± 25.65 & 22.3 ± 21.30 \\
\hline
\end{tabular}
\vspace{0.2cm}
\footnotesize
\begin{tabular}{@{}p{\textwidth}@{}}
\textit{Note:}
Cluttered represents the average of ECPM and PAM;\\ Multi-stage represents the average of BHB, BS, ECP, and MH;\\ Overall represents the average of all six tasks. \\
Task-level standard deviations (e.g., 0.84-2.04)\\ measure performance consistency under repeated trials of the same task.\\ Group-level standard deviations (e.g., 6.17-25.65)\\ reflect variation across different tasks. 
\end{tabular}
\end{table}

\subsection{Overall Effectiveness}

The comparison of the overall success rate across these six
tasks mentioned above between our model and the baselines
is presented in TABLE \ref{success}.

In cluttered environment tasks, DP and DP3 struggles because they are only capable of replicating actions in the scenario they have seen. If the positions and types of interfer ing clutter change, they often failed to recognize the correct target and operate the task well. But our novel PRISM model is able to pay attention to the target object amidst clutter, enabling successfully manipulation. PRISM significantly outperforms DP and DP3 in these tasks.

For multi-stage tasks, PRISM also demonstrates a marked advantage over DP3. This improvement stems from PRISM’s ability to independently process and emphasize distinct targets during different phases of manipulation. DP3 tends to processes the entire scene uniformly, often missing critical details crucial for task success. Although DP may perform adequately in simpler settings, it falls short in addressing the varied objectives typical of multi-stage operations.

TABLE \ref{success} clearly shows that PRISM’s overall average success rate of 82.0\% far exceeds the 58.4\% and 22.3\% achieved by DP3 and DP, respectively. These results indicate that our approach—combining segmented point cloud features with cross-attention and a diffusion-based decoder—delivers exceptionally accurate continuous action prediction. Compared to 3D policies that often depend on pre-trained point cloud encoders) or large-scale pre-training datasets, PRISM achieves competitive performance while being trained from scratch. In essence, PRISM’s design is highly effective, providing a straightforward and powerful solution for robot manipulation tasks.

\section{Conclusion}
In this paper, we introduce PRISM, an end-to-end frame- work that leverages 3D point cloud segmentation and cross- attention to fuse visual and proprioceptive data for robotic manipulation. PRISM starts by decomposing raw point clouds into object-centric clusters and then integrates these local features with joint state information through a cross-attention mechanism. These conditioned features are subsequently used by a diffusion module to generate smooth and continuous actions.

Experimental results on a variety of challenging tasks—including cluttered environment manipulation and multi-stage scenarios—demonstrate that PRISM significantly outperforms conventional 2D and 3D policies, both in terms of accuracy and efficiency. The simulation-validated capabilities reported here establish a foundational framework ready for future hardware iterations. Our ablation studies further highlight the critical contribution of the transformer encoder, while also revealing its potential drawbacks in low-data regimes.

Overall, PRISM delivers a robust, flexible pipeline for 3D perception in imitation learning. Its selective focusing mechanism, which isolates task-relevant clusters before action decoding, can also be adopted within Vision-Language-Action (VLA) models by using language embeddings to query these segmented clusters. This architecture inherently supports transitional validation pathways toward physical systems, where object-level grounding is greatly enhanced, paving the way for more precise, interpretable language-conditioned manipulation across simulated and physical domains.
\bibliographystyle{ieeetr}
\bibliography{IEEEabrv,references}

\end{document}